%% file: neurips_2026.tex
\documentclass{article}

\usepackage[dblblindworkshop, final]{neurips_2026}

\usepackage[utf8]{inputenc} % allow utf-8 input
\usepackage[T1]{fontenc}    % use 8-bit T1 fonts
\usepackage{hyperref}       % hyperlinks
\usepackage{url}            % simple URL typesetting
\usepackage{booktabs}       % professional-quality tables
\usepackage{amsfonts}       % blackboard math symbols
\usepackage{nicefrac}       % compact symbols for 1/2, etc.
\usepackage{microtype}      % microtypography
\usepackage{xcolor}         % colors

\usepackage{times}
\usepackage{latexsym}
\usepackage{amsmath}
\usepackage{amssymb}
\usepackage[T1]{fontenc}
\usepackage[utf8]{inputenc}

\usepackage{microtype}
\usepackage{booktabs}
\usepackage{inconsolata}
\usepackage{subcaption}
\usepackage{graphicx}
\usepackage[table]{xcolor}

\title{C$^3$: Speech-Guided KV Retrieval for Efficient Long-Context SpeechLLM Inference}

\author{%
  Dancheng Liu\thanks{Work done during an internship at Amazon. Correspondence to: Srikanth Ronanki <ronanks@amazon.com> and Aram Galstyan <argalsty@amazon.com>.} \\
  University at Buffalo, USA
  \And
  Markus Müller\\
  Amazon AGI, USA
  \And
  Rupak Vignesh Swaminathan\\
  Amazon AGI, USA
  \And
  Jing Liu\\
  Amazon AGI, USA
  \And
  Srikanth Ronanki\\
  Amazon AGI, USA
  \And
  Aram Galstyan\\
  Amazon AGI, USA
}

\newcommand{\name}[0]{\textsc{C}\textsuperscript{3}}

\begin{document}

\maketitle

\input{content/00_abstract}

\input{content/01_introduction}

\input{content/04_method}

\input{content/05_results}

\input{content/06_conclusion}

% \begin{ack}
% Use unnumbered first level headings for the acknowledgments. All acknowledgments
% go at the end of the paper before the list of references. Moreover, you are required to declare
% funding (financial activities supporting the submitted work) and competing interests (related financial activities outside the submitted work).
% More information about this disclosure can be found at: \url{https://neurips.cc/Conferences/2026/PaperInformation/FundingDisclosure}.

% Do {\bf not} include this section in the anonymized submission, only in the final paper. You can use the \texttt{ack} environment provided in the style file to automatically hide this section in the anonymized submission.
% \end{ack}
\bibliographystyle{acl_natbib}
\bibliography{custom}

%%%%%%%%%%%%%%%%%%%%%%%%%%%%%%%%%%%%%%%%%%%%%%%%%%%%%%%%%%%%
\appendix

\input{appendix/appendix}

\end{document}

%% file: content/00_abstract.tex
\begin{abstract}
Speech large language models encode audio at a much higher rate than text, causing the KV cache to become a major bottleneck in long-context inference. While KV retrieval reduces decoding cost by attending to only a query-dependent subset of the cache, existing methods are designed primarily for text. We show that retrieval errors can compound into generation drift, and identify three properties of audio attention that guide more reliable retrieval: unequal concentration across query heads, non-contiguous audio support, and temporal continuity. Based on these observations, we propose \name{} (\emph{Cluster--Compete--Compete}), which combines hierarchical key-space clustering, shared retrieval competition within KV groups, and adaptive cluster resolution. Across long-form recognition, translation, and summarization on four SpeechLLMs, \name{} achieves the strongest overall accuracy under matched retrieval budgets, maintains substantially higher accuracy at tight budgets than prior methods, and improves accelerator-memory scaling and decoding efficiency.
\end{abstract}

%% file: content/01_introduction.tex
\section{Introduction} 

Speech large language models (SpeechLLMs) integrate audio and text within a unified framework, enabling general-purpose speech understanding and generation directly from raw audio \citep{tang2024salmonn,xu2025qwen3}. Compared with text, audio is encoded at a much higher temporal resolution, causing the key-value (KV) cache to grow rapidly with input duration and imposing substantial memory and decoding costs for long-form inference \citep{lu2026compress,yang2026longspeech}. 

KV retrieval addresses this bottleneck by retaining the full cache while attending to only a small, query-dependent subset at each decoding step \citep{tang2024quest,lee2024infinigen}. However, existing methods are designed primarily for text and may not match the structure of audio attention. We find that retrieval errors can compound autoregressively into generation drift, and that audio attention exhibits three properties important for retrieval under tight budgets: query heads sharing a KV head differ substantially in attention concentration, relevant audio support is often non-contiguous, and high-attention regions evolve gradually across decoding steps. 

Guided by these observations, we introduce \name{} (\emph{Cluster--Compete--Compete}), a per-KV-head retrieval method for SpeechLLMs. \name{} organizes audio KVs into hierarchical key-space clusters, lets query heads sharing a KV head compete for a common retrieval budget, and dynamically reallocates cluster resolution through paired split--merge updates. This concentrates fine retrieval resolution around currently relevant regions while representing the remainder of the audio more coarsely. Across long-form recognition, translation, and summarization on four SpeechLLMs, \name{} achieves the strongest overall retrieval accuracy under matched budgets, with its largest gains under tight retrieval constraints. On Qwen3-Omni, \name{} reduces LS-Long WER at a 128-KV budget from 180.94\% with ClusterKV \citep{liu2024clusterkv} to 7.07\%, and improves speech translation at a 256-KV budget from 25.91 to 55.79 chrF. It also improves long-context memory scaling and decoding efficiency. Our contributions are: 

\begin{itemize} 
\item We identify retrieval-induced generation drift and three properties of audio attention that motivate speech-specific KV retrieval design. 
\item We propose \name{}, combining hierarchical key-space clustering, shared query-head competition, and query-adaptive resolution under a fixed retrieval budget. 
\item We demonstrate strong retrieval accuracy across four SpeechLLMs and multiple long-form speech tasks, together with improved memory scaling and decoding efficiency. 
\end{itemize}

%% file: content/04_method.tex
\section{C$^3$: \textit{Cluster--Compete--Compete}}
\label{sec:method}

\begin{figure*}[t]
\centering
\includegraphics[width=\linewidth]{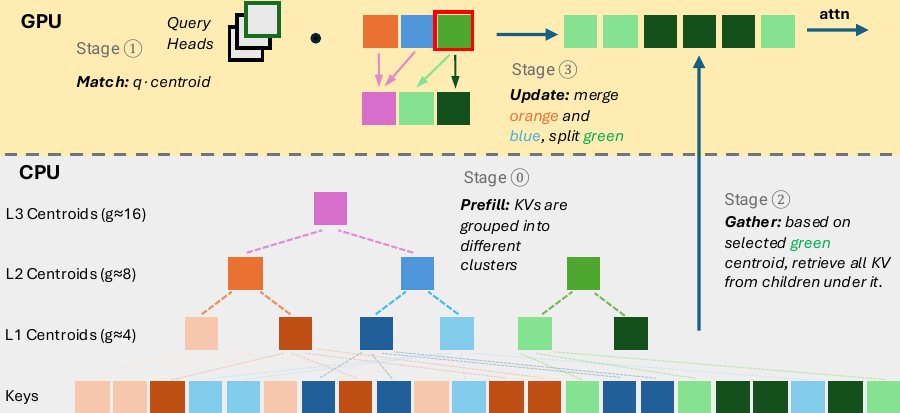}
\caption{Overview of \name{}. Audio KVs are organized into a hierarchical key-space index. Query heads sharing a KV head compete over an active frontier for a common retrieval budget, while paired split--merge operations dynamically reallocate frontier resolution.}
\vspace{-5mm}
\label{fig:method_overview}
\end{figure*}

Motivated by the audio-attention properties analyzed in Appendix~\ref{app:analysis}, we introduce \name{} (\emph{Cluster--Compete--Compete}), a per-KV-head retrieval method for SpeechLLMs. For each layer and KV head, \name{} constructs a hierarchical key-space index over the audio KVs during prefill. During decoding, query heads sharing a KV head compete for a common retrieval budget, while the hierarchy dynamically reallocates resolution toward regions relevant to recent queries. Figure~\ref{fig:method_overview} summarizes the procedure. System implementation and kernel optimizations are described in Appendix~\ref{sec:hw_appendix}.

\paragraph{Stage 0: Hierarchical prefilling.}
For each KV head, \name{} recursively bisects the audio keys using k-means to form a binary hierarchy. Each node represents the keys beneath it using its centroid. At the beginning of decoding, $A$ nodes from an intermediate level form an \emph{active frontier}, a partition of the audio keys in which every token belongs to exactly one active node. The frontier size remains fixed, but its nodes may occupy different hierarchy depths, allowing relevant regions to be represented more finely than inactive regions.

\paragraph{Stage 1: Compete for the retrieval budget.}
Under grouped-query attention, let $G$ query heads share one KV head. At decoding step $t$, each active node $n$ is scored by its strongest cosine-similarity match with any associated query head:
\begin{equation}
\sigma_{t,n}
=
\max_{g\in[G]}
\widehat{\mathbf{q}}_{t}^{(g)\top}
\widehat{\mathbf{c}}_n ,
\label{eq:node_score}
\end{equation}
where $\widehat{\mathbf{q}}_{t}^{(g)}$ and $\widehat{\mathbf{c}}_n$ are $\ell_2$-normalized queries and centroids. Taking the maximum allows any strongly matching query head to promote a cluster without averaging away concentrated attention. The $G$ query heads therefore produce a single cluster ranking and compete for one common KV budget rather than retrieving independent subsets. %Alternative query-head aggregation strategies are evaluated in Appendix~\ref{app:ablations}.

\paragraph{Stage 2: Gather the winning KVs.}
Active nodes are consumed in descending order of $\sigma_{t,n}$ until the per-KV-head retrieval budget $B$ is filled. The corresponding audio keys and values are gathered into a bounded retrieval cache, and all $G$ query heads compute exact attention over the same subset of at most $B$ audio KVs.

\paragraph{Stage 3: Compete for resolution.}
To adapt retrieval granularity over generation, \name{} maintains an exponential moving average of each active node's relevance:
\begin{equation}
e_{t,n}
=
\lambda e_{t-1,n}
+
(1-\lambda)\sigma_{t,n}.
\label{eq:ema_score}
\end{equation}
Nodes that remain strongly matched to recent queries receive higher refinement priority. At each step, \name{} replaces a high-priority active node with its two children, exposing finer retrieval units in that region. To keep the frontier size fixed, a low-priority active sibling pair is simultaneously replaced by its parent. These paired split--merge operations therefore move retrieval resolution toward currently relevant regions without increasing the number of active representatives. Newly activated nodes inherit the recent relevance state of the nodes they replace. %Hyperparameter and adaptation ablations are provided in Appendix~\ref{app:ablations}.

\paragraph{Computation.}
Let $N$ denote the number of audio KVs, $A$ the active-frontier size, and $B$ the retrieval budget. Full attention scores all $N$ audio KVs, whereas \name{} scores only $A$ centroids and computes exact attention over at most $B$ retrieved KVs. With initial cluster granularity $g$, $A\approx N/g$, while final-KV retrieval and exact-attention computation remain bounded by $B$.%Detailed system costs and host-offloaded execution are described in Appendix~\ref{sec:hw_appendix}.

%% file: content/05_results.tex
\section{Experiments and Results}
\label{sec:experiments}

\paragraph{Experimental setup.}
We evaluate \name{} on four SpeechLLMs: Qwen3-Omni-30B-A3B \citep{xu2025qwen3}, Qwen2.5-Omni-3B \citep{Qwen2.5-Omni}, Gemma-4-E4B \citep{team2026gemma}, and Voxtral-Mini-3B \citep{liu2025voxtral}, using a single NVIDIA H200 GPU. Efficiency evaluations also use an Ada 6000 environment, matching the setup from \citet{liu2024clusterkv}. Our main evaluation covers long-form ASR on LibriSpeech-Long (LS-Long) \citep{park2024long} and a multilingual benchmark constructed from Multilingual LibriSpeech (MLS) \citep{pratap2020mls}, speech translation and summarization on LongSpeech \citep{yang2026longspeech}. We report WER percentage ($\downarrow$) for ASR, chrF ($\uparrow$) for translation, and ROUGE-L ($\uparrow$) for summarization. We denote content-based clustering as C1, shared query-head competition as C2, and adaptive resolution as C3; the complete C1+C2+C3 configuration is \name{}. We compare against full attention, Quest \citep{tang2024quest}, UNIQUE \citep{deng2026unique}, and ClusterKV \citep{liu2024clusterkv} under matched final audio-KV retrieval budgets. Dataset construction, inference settings, budget matching, and method hyperparameters are detailed in Appendix~\ref{sec:exp_details}.

\input{table/main_table}

\paragraph{Long-form speech accuracy.}
Table~\ref{tab:main_table} shows that \name{} achieves the strongest overall retrieval performance across models, tasks, and budgets, with its largest gains under tight retrieval constraints. The complete configuration is the best retrieval method in 31 of the 32 settings shown. At larger budgets, the gap to full attention narrows substantially in most settings, while \name{} remains markedly more robust than prior retrieval methods at tight budgets. Complete budget sweeps, component and hyperparameter ablations, and additional audio-understanding results are provided in Appendix~\ref{app:full_results}--\ref{app:audio_understanding}.

\begin{figure*}[t]
    \centering

    \begin{minipage}[t]{0.49\textwidth}
        \centering
        \includegraphics[height=3cm]
        {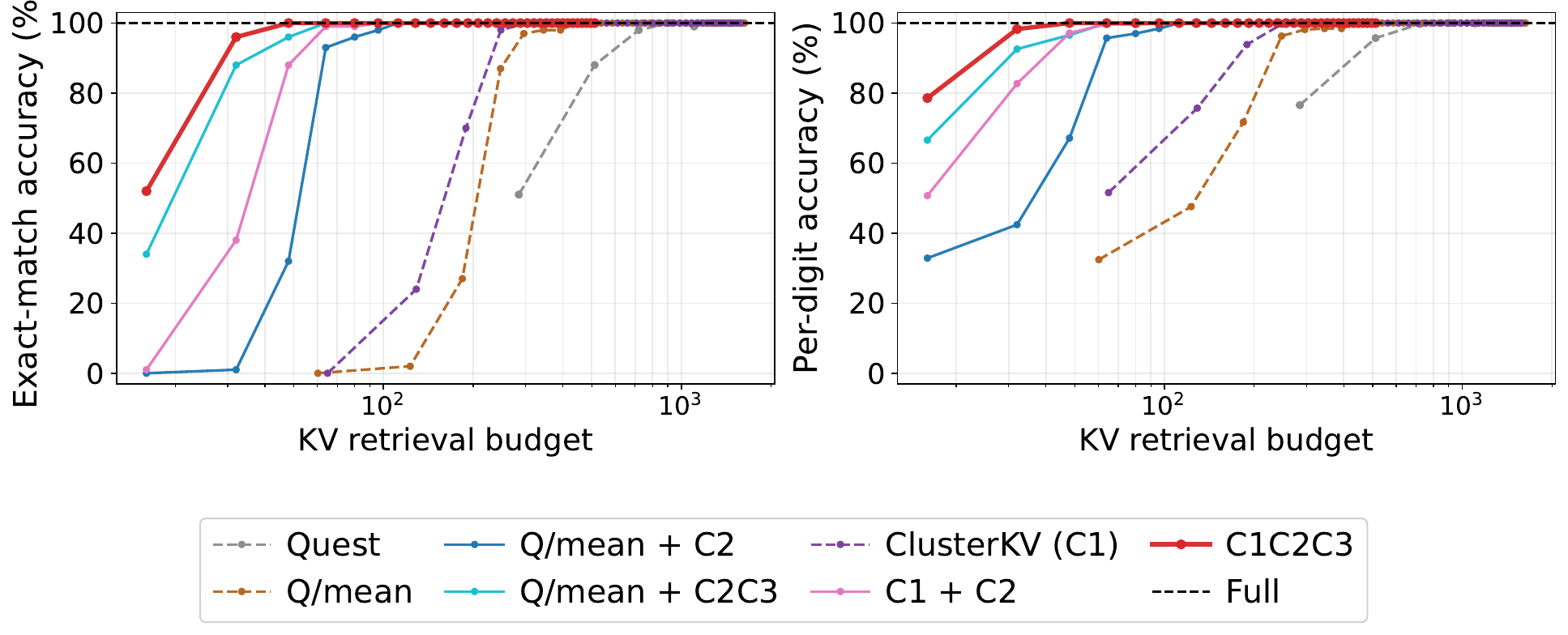}
        \vspace{-2mm}

        \textbf{(a) Needle-in-a-haystack retrieval}
    \end{minipage}
    \hfill
    \begin{minipage}[t]{0.49\textwidth}
        \centering
        \includegraphics[height=3cm]
        {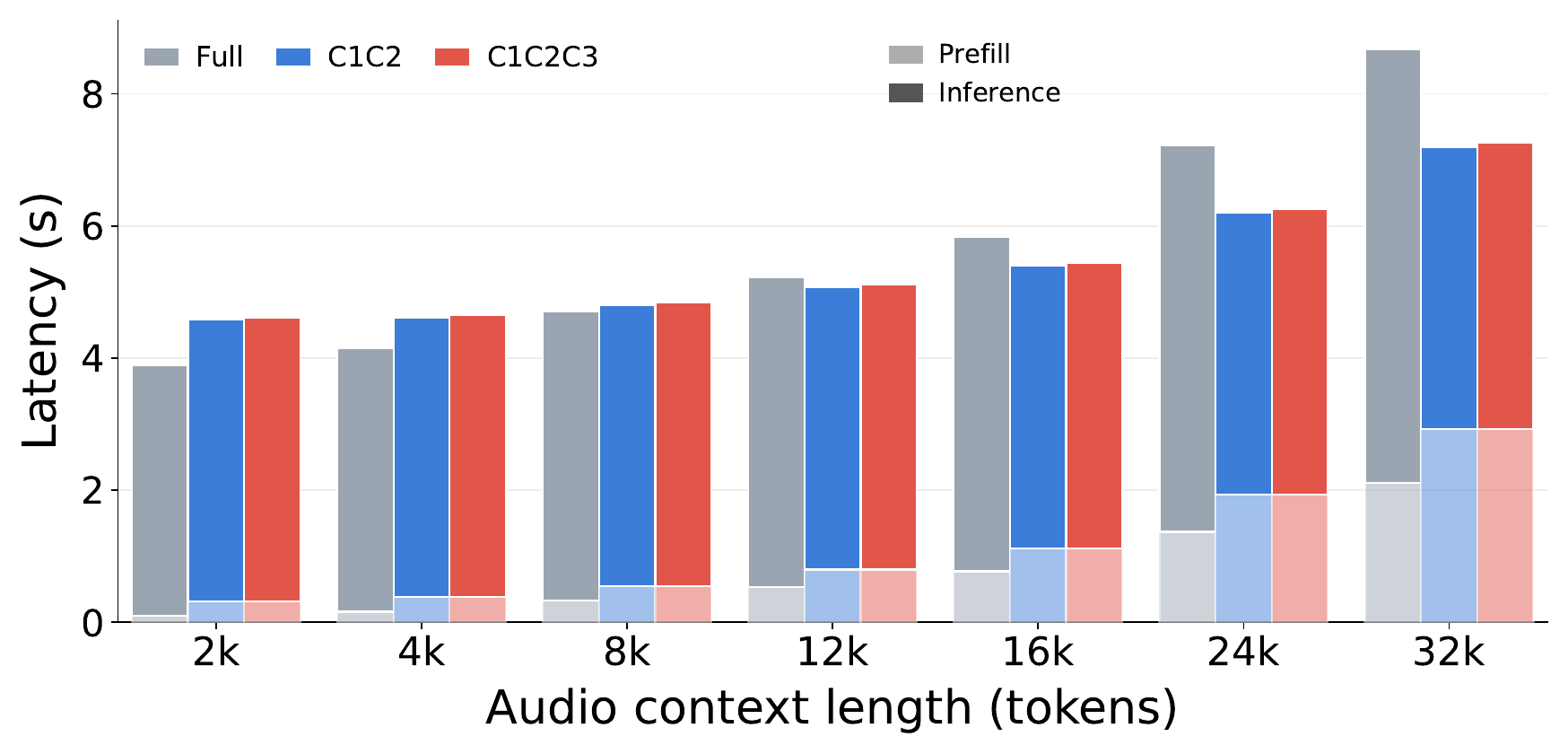}
        \vspace{-2mm}

        \textbf{(b) End-to-end inference latency}
    \end{minipage}

    \caption{
        % Retrieval quality and efficiency on Qwen3-Omni-30B-A3B.
        \textbf{(a)} \name{} achieves lossless passcode recovery at
        substantially smaller retrieval budgets than the baselines.
        \textbf{(b)} End-to-end latency for audio contexts up to 32K
        tokens with a 512-KV retrieval budget; retrieval becomes more
        efficient than full attention as the audio context grows.
    }
    \label{fig:niah_efficiency}
    \vspace{-5mm}
\end{figure*}

\paragraph{Needle-in-a-haystack retrieval.}
We further test whether retrieval preserves isolated evidence using 120 approximately 13--16 minute recordings containing three spoken seven-digit passcodes inserted at separate locations. On Qwen3-Omni, Figure~\ref{fig:niah_efficiency}(a) shows that \name{} reaches 95\% exact-match accuracy with only 32 retrieved KVs and lossless recovery with 64 KVs, whereas ClusterKV and temporal baselines require several hundred KVs to approach the same accuracy. Full benchmark construction, per-digit accuracy, and results on the other models are provided in Appendix~\ref{sec:app_niah}.

\paragraph{Memory and decoding efficiency.}
Figure~\ref{fig:niah_efficiency}(b) reports end-to-end latency on Qwen3-Omni-30B-A3B for audio contexts up to 32K tokens using a 512-KV retrieval budget. Retrieval introduces additional scoring, selection, and gathering overhead at short contexts, but both C1+C2 and the complete \name{} become faster than full attention at approximately 8K audio tokens. Adaptive resolution adds only modest overhead while retaining the tight-budget accuracy gains in Table~\ref{tab:main_table}. For longer contexts, the full audio KV bank and inactive hierarchy nodes can additionally be offloaded to host memory, reducing accelerator-resident audio state from $O(N)$ to $O(A+B)$. %Full profiling, memory, throughput, and host-offloading results are provided in Appendix~\ref{app:eff_discussion}.

%% file: table/main_table.tex
% requires: \usepackage{booktabs}, \usepackage[table]{xcolor}
\begin{table*}[t]
    \centering
    \caption{Long-form speech performance under tight and large audio KV
    retrieval budgets. Best retrieval results within each model and budget are shown in bold. Results at intermediate budgets and additional ablations are provided in Appendix~\ref{app:full_results}.} %We report WER ($\downarrow$) for LibriSpeech-Long (LS-Long) and multilingual long-form ASR (MLS), chrF ($\uparrow$) for speech translation (ST), and ROUGE-L ($\uparrow$) for summarization (Summ). Full attention is included as a reference. Best retrieval results within each model and budget are shown in bold, and the complete \name configuration is shaded. A dash (--) indicates that Quest does not support the target budget. Results at intermediate budgets and additional ablations are provided in Appendix~\ref{app:full_results}.} 
    \label{tab:main_table}

    \setlength{\tabcolsep}{8.0pt}
    \small
    \renewcommand{\arraystretch}{0.99}
\begin{tabular}{l cc cc cc cc}
        \toprule
        & \multicolumn{2}{c}{LS-Long($\downarrow$)}
        & \multicolumn{2}{c}{MLS($\downarrow$)}
        & \multicolumn{2}{c}{ST($\uparrow$)}
        & \multicolumn{2}{c}{Summ($\uparrow$)} \\
        \cmidrule(lr){2-3}
        \cmidrule(lr){4-5}
        \cmidrule(lr){6-7}
        \cmidrule(lr){8-9}
        KV Budget
        & 128 & 512
        & 256 & 1024
        & 256 & 1024
        & 256 & 1024 \\
        \midrule
\multicolumn{9}{c}{\textbf{Qwen3 Omni 30B A3B}} \\
\midrule
Full & \multicolumn{2}{c}{2.02} & \multicolumn{2}{c}{4.83} & \multicolumn{2}{c}{74.75} & \multicolumn{2}{c}{25.52} \\
\midrule
Quest & -- & 193.28 & 131.87 & 119.96 & 5.48 & 11.52 & 18.80 & 22.01 \\
UNIQUE & 70.30 & 2.53 & 45.20 & 5.39 & 21.15 & 63.68 & 22.31 & 24.22 \\
C1 (ClusterKV) & 180.94 & 2.81 & 75.00 & 5.62 & 25.91 & 70.15 & 24.06 & 24.91 \\
\rowcolor{gray!15}
\quad +C2+C3 & \textbf{7.07} & \textbf{1.98} & \textbf{10.05} & \textbf{5.18} & \textbf{55.79} & \textbf{71.96} & \textbf{24.71} & \textbf{25.29} \\
\midrule
\multicolumn{9}{c}{\textbf{Qwen2.5 Omni 3B}} \\
\midrule
Full & \multicolumn{2}{c}{2.40} & \multicolumn{2}{c}{14.52} & \multicolumn{2}{c}{42.13} & \multicolumn{2}{c}{24.01} \\
\midrule
Quest & 142.84 & 126.52 & 155.94 & 116.72 & 7.47 & 16.53 & 18.64 & 21.72 \\
UNIQUE & 51.49 & 4.55 & 43.83 & 15.98 & 15.49 & 34.20 & 21.24 & 23.36 \\
C1 (ClusterKV) & 133.13 & 3.91 & 65.32 & 15.61 & 19.32 & 35.01 & 22.68 & 23.97 \\
\rowcolor{gray!15}
\quad +C2+C3 & \textbf{40.48} & \textbf{2.78} & \textbf{31.98} & \textbf{14.31} & \textbf{23.76} & \textbf{39.35} & \textbf{22.71} & \textbf{24.11} \\
\midrule
\multicolumn{9}{c}{\textbf{Gemma-4-E4B}} \\
\midrule
Full & \multicolumn{2}{c}{4.24} & \multicolumn{2}{c}{10.26} & \multicolumn{2}{c}{58.97} & \multicolumn{2}{c}{21.64} \\
\midrule
Quest & -- & 122.78 & 110.26 & 90.70 & 2.35 & 6.73 & 17.55 & 19.89 \\
UNIQUE & 42.77 & 6.76 & 34.09 & 12.72 & 26.03 & 47.19 & 18.91 & 19.59 \\
C1 (ClusterKV) & 95.72 & 16.73 & 64.05 & 19.26 & 21.28 & 42.17 & \textbf{20.19} & 20.72 \\
\rowcolor{gray!15}
\quad +C2+C3 & \textbf{14.63} & \textbf{5.43} & \textbf{18.44} & \textbf{10.98} & \textbf{42.29} & \textbf{54.93} & 19.83 & \textbf{20.97} \\
\midrule
\multicolumn{9}{c}{\textbf{Voxtral Mini 3B}} \\
\midrule
Full & \multicolumn{2}{c}{3.39} & \multicolumn{2}{c}{7.29} & \multicolumn{2}{c}{68.43} & \multicolumn{2}{c}{26.94} \\
\midrule
Quest & -- & 164.59 & 190.96 & 136.27 & 12.21 & 10.14 & 16.91 & 20.12 \\
UNIQUE & 104.74 & 10.66 & 58.48 & 8.58 & 5.00 & 29.17 & 19.35 & 22.40 \\
C1 (ClusterKV) & 212.96 & 33.64 & 107.58 & 8.21 & 12.42 & 43.74 & 23.29 & 25.74 \\
\rowcolor{gray!15}
\quad +C2+C3 & \textbf{41.86} & \textbf{4.61} & \textbf{14.65} & \textbf{7.61} & \textbf{23.02} & \textbf{48.84} & \textbf{24.35} & \textbf{25.84} \\
\bottomrule
\end{tabular}
% \vspace{-3mm}
\end{table*}

%% file: content/06_conclusion.tex
\section{Conclusion}

In this work, we study audio KV retrieval for long-form SpeechLLMs and show how retrieval errors can compound autoregressively into generation drift. Motivated by unequal query-head concentration, non-contiguous audio support, and temporal continuity, we introduced \name{}, which combines hierarchical key-space clustering, shared competition within KV groups, and adaptive split--merge refinement. Across four SpeechLLMs and multiple long-form speech tasks, \name{} achieves the strongest overall retrieval accuracy under tight budgets while improving long-context memory scaling and decoding efficiency. %These results demonstrate the importance of designing KV retrieval around the structure of audio attention.

%% file: appendix/appendix.tex
\input{appendix/sections/secA}
\input{appendix/sections/secB}
\input{appendix/sections/secC}
\input{appendix/sections/secD}

\input{appendix/sections/secE}

%% file: appendix/sections/secA.tex
\section{Limitations}
\label{sec:limitations}

Our evaluation covers four SpeechLLMs and multiple long-form speech
tasks, but the observed accuracy and efficiency trade-offs may vary
with other model architectures, audio encoders, and generation
settings. In particular, the benefits of KV retrieval depend on how
much of a model uses token-level full attention. For example, hybrid architectures
with fewer full-attention layers naturally expose less KV-cache state
to optimize.

System performance also depends on hardware and implementation.
Our measurements use a single NVIDIA H200, together with Ada 6000 GPUs when evaluating efficiency. The context length at which retrieval becomes faster than full
attention may therefore shift on other accelerators or with more
specialized kernels. Similarly, under fixed cluster granularity, the
number of active representatives grows with audio length, so extremely
long contexts may require coarser granularity or an additional
hierarchical search stage.

Finally, end-to-end model evaluations are limited to the context
lengths supported by the SpeechLLMs studied here. System results beyond
32K audio tokens are analytical projections and demonstrate expected
memory and throughput scaling rather than generation quality at those
lengths. Evaluating retrieval on future SpeechLLMs with substantially
longer native audio contexts is therefore an important direction for
future work.

\section{Broader Impacts}
\label{sec:broader_impacts}

Improving KV-cache efficiency can make long-form SpeechLLMs more
practical on memory- and latency-constrained hardware, potentially
reducing deployment cost and energy use. More efficient long-context
speech processing may also broaden access to applications such as
transcription, translation, and audio understanding. At the same time,
greater efficiency can lower the cost of deploying SpeechLLMs in
settings involving sensitive audio, where privacy, surveillance, and
misuse risks depend primarily on the underlying model and application.
These risks should therefore be considered when deploying systems that
incorporate the proposed retrieval method.

\section{Related Work}
\label{sec:related_ext}

KV-cache methods reduce inference cost through selection
\citep{tang2024quest,zhang2023h2o}, quantization
\citep{liu2024kivi,hooper2024kvquant,zhang2024kv}, or architectural
changes such as grouped-query attention, multi-head latent attention,
and state-space layers
\citep{ainslie2023gqa,liu2024deepseek,gu2024mamba}. These directions
are complementary. We focus on selection, where eviction permanently
discards cached entries, whereas retrieval preserves the backing cache
and dynamically restricts the entries accessed by each query.

\paragraph{Eviction.}
Eviction retains entries according to signals such as recency and
attention sinks \citep{xiao2024efficient}, accumulated attention
\citep{zhang2023h2o,oren2024transformers,lin2025compresskv}, or
prompt-based importance \citep{li2024snapkv}. Adaptive methods further
vary budgets or policies across heads and layers
\citep{ge2023model,cai2024pyramidkv,feng2026ada}. Because discarded
entries cannot be recovered, eviction can be problematic for speech
generation, where an audio frame that appears unimportant early may become
necessary when its corresponding content is generated. AudioKV
mitigates this issue by allocating larger budgets to audio-critical
heads and smoothing importance estimates across neighboring frames,
but still requires approximately 40\% KV retention for near-lossless
performance \citep{wang2026audiokv}.

\paragraph{Retrieval.}
Retrieval preserves the full cache while selecting a query-dependent
subset at each decoding step
\citep{tang2024quest,sun2024shadowkv}. Existing methods retrieve fixed
positional pages or chunks
\citep{tang2024quest,sun2024shadowkv,deng2026unique}, similarity-based
key-space clusters \citep{liu2024clusterkv}, or candidates obtained
through coarse-to-fine search
\citep{hooper2025squeezed,hooper2026multipole,li2026lycheecluster}.
Related approaches accelerate candidate identification using
attention-aware approximate nearest-neighbor indexes, product
quantization, or locality-sensitive hashing
\citep{liu2026retrievalattention,zhang2025pqcache,chen2025magicpig}.
Most of these methods were developed for text and use fixed retrieval
units or independently recompute their selection at each step.
\name{} instead combines non-contiguous key-space units, shared
competition among query heads associated with the same KV head, and
persistent split--merge updates that adapt retrieval resolution as
audio relevance evolves.

\paragraph{Speech-specific methods.}
Recent work exploits modality-specific structure in SpeechLLM caches.
SpeechKV learns to pool acoustically redundant speech KVs
\citep{lu2026compress}; Murmur combines chunked ASR inference with
sliding-window eviction \citep{lee2026murmur}; and TRADE bounds cache
growth using localized decoder audio attention
\citep{tang2026trade}. Together with AudioKV, these methods reduce the
retained or accessible context through eviction, aggregation, or
architecturally constrained attention. They therefore address a
different operating point from query-dependent retrieval.

Concurrent work UNIQUE applies training-free page-based sparse
attention to text and speech models \citep{deng2026unique}. It scores
fixed contiguous pages using mean keys and a dispersion correction,
with page selection shared by query heads associated with the same KV
head. This structure is closest to our \textit{Quest/mean+C2}
ablation, which does not include the dispersion correction. In
contrast, \name{} constructs non-contiguous units in key space and
persistently reallocates their resolution through paired split--merge
updates, allowing the retrieval structure to follow changes in audio
attention throughout generation.

%% file: appendix/sections/secB.tex
\section{Speech-Guided KV Retrieval Analysis}
\label{app:analysis}

This section of the appendix analyzes retrieval-induced generation drift and three
properties of audio attention that motivate \name{}: unequal
concentration across query heads, non-contiguous audio support, and
temporal continuity across decoding steps.

\subsection{Failure Mode: Retrieval-Induced Generation Drift}
\label{sec:ana_failure}

\begin{figure}[t]
\centering
\includegraphics[width=\linewidth]{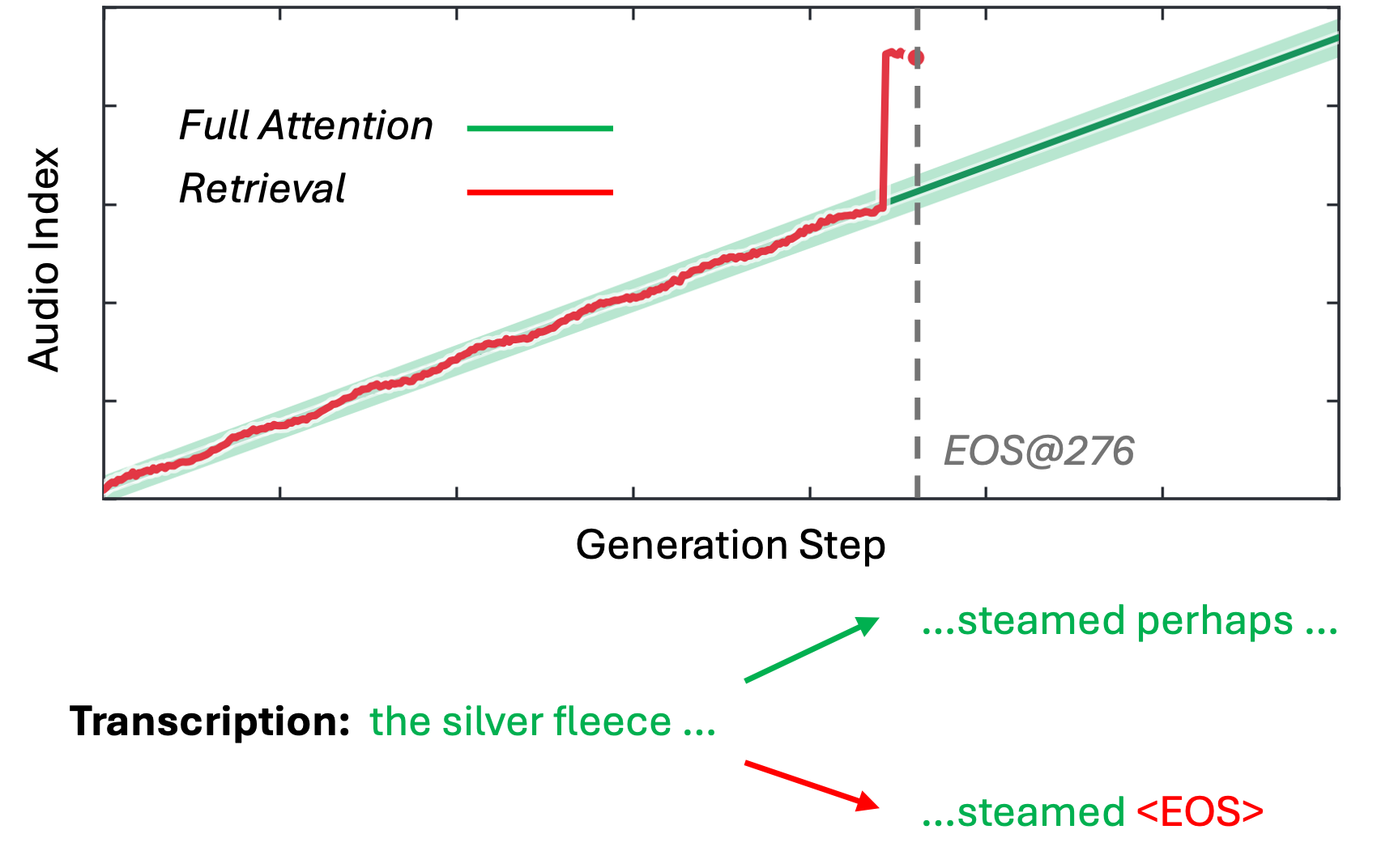}
\caption{Retrieval-induced generation drift. The retrieval trajectory
initially follows the relevant audio region but later jumps forward,
causing the transcript to lose alignment and terminate prematurely.
Curves are smoothed and outliers removed for visualization.}
\label{fig:retrieval_failure}
\end{figure}

KV retrieval can progressively misalign the decoder state from the
input audio. In the ASR example in
Figure~\ref{fig:retrieval_failure}, the retrieved region initially
advances with the transcript but jumps to a later audio segment at
decoding step 276. The generated transcript subsequently loses
alignment with the speech and emits an end-of-sequence token
prematurely.

Let $\mathcal{R}_t$ contain the non-audio context (which is always retained) and retrieved audio
positions at decoding step $t$. Their retained mass under full
attention is
\begin{equation}
\rho_t
=
\sum_{i\in\mathcal{R}_t}\alpha_{t,i},
\label{eq:retained_mass}
\end{equation}
where $\alpha_{t,i}$ is the full-attention weight of position $i$.
As derived in Appendix~\ref{sec:drift_derivation}, retrieved and full
attention for the same query differ by
\begin{equation}
\widetilde{\mathbf{o}}_t-\mathbf{o}_t
=
(1-\rho_t)\mathbf{d}_t,
\label{eq:retrieval_drift}
\end{equation}
where $\mathbf{d}_t$ is the discrepancy between the
attention-weighted value representations of the retained and omitted
positions. The resulting value-agnostic bound is
\begin{equation}
\left|
\widetilde{\mathbf{o}}_t-\mathbf{o}_t
\right|_2
\leq
(1-\rho_t)D_t,
\label{eq:retrieval_drift_bound}
\end{equation}
where $D_t$ is the diameter of the value vectors. Thus, the
perturbation depends on both the omitted attention mass
$1-\rho_t$ and the difference between the retained and omitted
values.

Although one perturbation need not change the generated token, its
effect can compound autoregressively. The altered attention output
changes the hidden state and next query, potentially changing the next
retrieval decision and further separating generation from the
full-attention trajectory. Repetition of this feedback loop can eventually produce
the drift observed in Figure~\ref{fig:retrieval_failure}.

Under a budget of $K$ audio positions, exact full-attention top-$K$
selection maximizes $\rho_t$ and minimizes the bound in
Equation~\ref{eq:retrieval_drift_bound}. This oracle requires full
attention, however. Practical methods must approximate it using compact
group-level representatives, which may retrieve low-relevance members
alongside relevant ones or omit relevant positions assigned to other
groups. Reliable approximation therefore depends on how retrieval is
coordinated across query heads, how audio KVs are grouped, and at what
resolution the groups are represented.

\subsection{Shared Retrieval Across Query Heads}
\label{sec:ana_head}

\begin{figure}[t]
\centering
\includegraphics[width=\columnwidth]{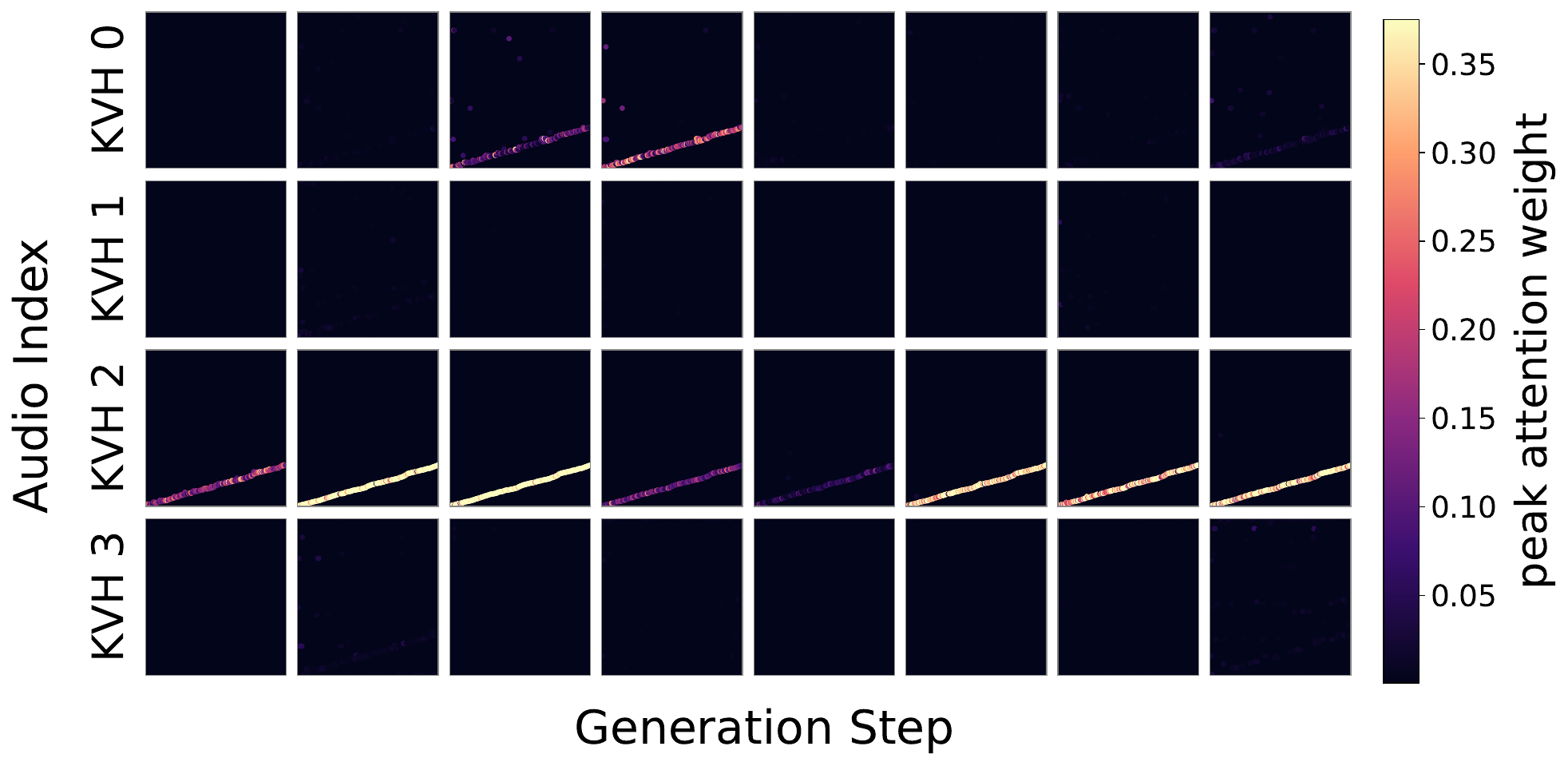}
\caption{Attention specialization across KV and query heads. Query
heads tracking task-relevant acoustic information concentrate on
similar regions within each KV group, whereas heads with different
preferences exhibit more diffuse attention.}
\label{fig:qhead}
\end{figure}

Audio relevance differs across KV heads, where some follow the advancing
speech frontier, whereas others repeatedly attend to persistent
acoustic anchors \citep{wang2026audiokv}. Retrieval decisions should
therefore remain distinct across KV heads. Within each KV group,
however, its associated query heads need not receive independent
budgets. Figure~\ref{fig:qhead} shows that concentrated query heads typically
focus on a few overlapping audio regions, whereas other heads
distribute their attention more diffusely. At the displayed layer, the
concentrated group contains roughly one third of the query heads and
has mean normalized attention entropy of approximately $0.40$, compared
with $0.79$ for the diffuse group. Concentrated heads place an average
of 85\% of their attention mass on audio, while diffuse heads place
only about 12\%. Within a KV group, concentrated heads also tend to
select overlapping top positions, whereas diffuse heads contribute
relatively little additional consensus.

Under a limited budget, omitting one highly weighted position can
remove substantial mass from a concentrated head, while omitting one
position has a smaller effect on a diffuse head. Independent,
comparable budgets for every query head may therefore spend excessive
capacity on diffuse patterns instead of prioritizing dominant
positions. Sharing a budget among query heads associated with the same
KV head allows the strongest concentrated matches to drive retrieval.
This motivates the max-based competition in
Equation~\ref{eq:node_score}, while preserving separate decisions
across KV heads.

\subsection{Non-Contiguity of Audio Relevance}
\label{sec:ana_band}

\begin{figure}[t]
\centering
\includegraphics[width=\columnwidth]{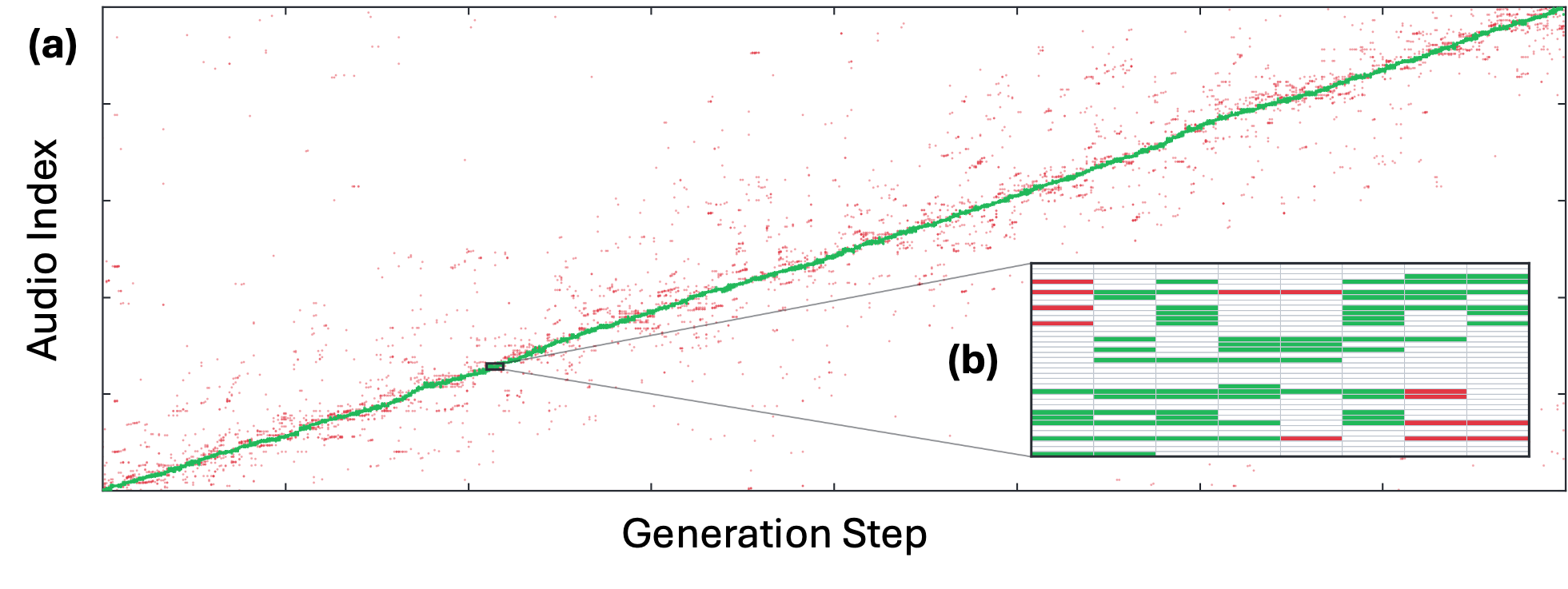}
\caption{Full audio attention during long-form decoding.
\textbf{(a)} The dominant attention band advances with generation;
red points mark top-$64$ positions outside the band.
\textbf{(b)} Attention near the frontier is itself non-contiguous.}
\label{fig:kv_structure}
\end{figure}

Audio attention is temporally structured but not confined to one
continuous interval. Figure~\ref{fig:kv_structure}(a) shows a dominant
attention frontier advancing through the audio during ASR decoding,
consistent with the alignment between generated text and its
corresponding audio region observed by
\citet{wang2026audiokv}. Nevertheless, the fraction of top-$64$
attention mass outside the dominant band averages 34\% for
recognition, 29\% for translation, and 78\% for summarization.
Restricting retrieval to the dominant temporal region would therefore
omit substantial highly weighted support.

The local region is also fragmented:
Figure~\ref{fig:kv_structure}(b) shows attention concentrated on
several separated positions rather than every frame in a continuous
interval. Temporally adjacent KVs consequently do not always form a
natural retrieval unit.

Fixed temporal chunks are therefore inefficient under a limited
budget. A chunk covering the dominant frontier may omit separated
relevant positions, while retrieving additional chunks includes the
low-relevance KVs between them. Key-space clustering better reflects
the attention mechanism: keys with similar representations tend to
receive similar scores for a query even when temporally distant,
allowing positions with similar query responses to share a retrieval
unit. This motivates the non-contiguous key-space clustering in
\name{}.

\subsection{Temporal Continuity of Audio Relevance}
\label{sec:ana_frontier}

Group-level retrieval trades retrieval precision against the number of
representatives scored. Smaller groups avoid retrieving low-relevance
KVs with relevant ones but require more representatives. Maintaining
fine resolution throughout the cache is therefore unnecessary if the
regions requiring such resolution can be anticipated from recent
decoding steps.

Audio attention provides this signal because high-attention regions
typically evolve gradually. In ASR, the dominant frontier progresses
approximately monotonically with the transcript; more generally, a
region that becomes relevant often remains relevant for several
subsequent steps. For decoding-step offsets
$\Delta\in{1,2,4,8}$, the mean Jaccard overlaps between top-$64$
attention sets are respectively
$0.58,0.46,0.34,0.24$ for recognition;
$0.56,0.44,0.33,0.22$ for translation; and
$0.45,0.34,0.24,0.16$ for summarization. On the ASR traces,
the peak position of audio-relevant heads moves by a median of
approximately four audio frames, or $0.3$s, per decoding step.

Fine retrieval resolution is thus temporally localized rather than
changing independently at every token. \name{} exploits this continuity
by concentrating fine groups around currently relevant regions and
representing inactive regions more coarsely. Fine groups improve
retrieval precision where needed, while coarse groups reduce the
number of representatives maintained and scored elsewhere. The
allocation can then follow the gradually evolving attention pattern
throughout generation.

\subsection{Derivation of Retrieval-Induced Attention Drift}
\label{sec:drift_derivation}

We provide the full derivation of
Equation~\ref{eq:retrieval_drift} and the corresponding bound in
Equation~\ref{eq:retrieval_drift_bound}.

Let $\mathcal{A}$ denote the set of audio-token positions and let
$\mathcal{C}_t$ denote the non-audio context positions available at
decoding step $t$, including prompt and previously generated tokens.
The full context is
$\mathcal{U}_t=\mathcal{A}\cup\mathcal{C}_t$.
For each position $i\in\mathcal{U}_t$, let $\mathbf{k}_i$ and
$\mathbf{v}_i$ denote its key and value. Full attention computes
\begin{equation}
\alpha_{t,i}
=
\frac{
e^{\mathbf{q}_t^\top\mathbf{k}_i/\sqrt{d}}
}{
\sum_{j\in\mathcal{U}_t}
e^{\mathbf{q}_t^\top\mathbf{k}_j/\sqrt{d}}
},
\end{equation}
and
\begin{equation}
\mathbf{o}_t
=
\sum_{i\in\mathcal{U}_t}
\alpha_{t,i}\mathbf{v}_i.
\end{equation}

Suppose retrieval selects an audio subset
$\mathcal{S}_t\subseteq\mathcal{A}$. Because non-audio positions remain
on the full-attention path, the complete retained set is
\begin{equation}
\mathcal{R}_t
=
\mathcal{C}_t\cup\mathcal{S}_t.
\end{equation}
Its attention mass under full attention is
\begin{equation}
\rho_t
=
\sum_{i\in\mathcal{R}_t}
\alpha_{t,i}
=
1-
\sum_{i\in\mathcal{A}\setminus\mathcal{S}_t}
\alpha_{t,i}.
\label{eq:app_retained_mass}
\end{equation}

We compare full and retrieved attention using the same query
$\mathbf{q}_t$. This isolates the instantaneous perturbation caused by
subset retrieval; its autoregressive propagation is discussed in
Section~\ref{sec:ana_failure}. Because retrieval leaves the query and
retained keys unchanged, it changes only the softmax normalization.
For $i\in\mathcal{R}_t$, the retrieved attention weight is
\begin{equation}
\widetilde{\alpha}_{t,i}
=
\frac{\alpha_{t,i}}{\rho_t}.
\end{equation}
Hence,
\begin{equation}
\widetilde{\mathbf{o}}_t
=
\frac{1}{\rho_t}
\sum_{i\in\mathcal{R}_t}
\alpha_{t,i}\mathbf{v}_i.
\label{eq:app_subset_attention}
\end{equation}

For $0<\rho_t<1$, the full-attention output can be written as
\begin{equation}
\mathbf{o}_t
=
\sum_{i\in\mathcal{R}_t}
\alpha_{t,i}\mathbf{v}_i
+
\sum_{i\notin\mathcal{R}_t}
\alpha_{t,i}\mathbf{v}_i.
\end{equation}
Subtracting this expression from
Equation~\ref{eq:app_subset_attention} gives
\begin{align}
\widetilde{\mathbf{o}}_t-\mathbf{o}_t
&=
\frac{1-\rho_t}{\rho_t}
\sum_{i\in\mathcal{R}_t}
\alpha_{t,i}\mathbf{v}_i
-
\sum_{i\notin\mathcal{R}_t}
\alpha_{t,i}\mathbf{v}_i
\nonumber\\
&=
(1-\rho_t)
\left(
\frac{1}{\rho_t}
\sum_{i\in\mathcal{R}_t}
\alpha_{t,i}\mathbf{v}_i
-
\frac{1}{1-\rho_t}
\sum_{i\notin\mathcal{R}_t}
\alpha_{t,i}\mathbf{v}_i
\right).
\label{eq:app_drift}
\end{align}

Defining
\begin{equation}
\mathbf{d}_t
=
\frac{1}{\rho_t}
\sum_{i\in\mathcal{R}_t}
\alpha_{t,i}\mathbf{v}_i
-
\frac{1}{1-\rho_t}
\sum_{i\notin\mathcal{R}_t}
\alpha_{t,i}\mathbf{v}_i,
\end{equation}
we obtain
\begin{equation}
\widetilde{\mathbf{o}}_t-\mathbf{o}_t
=
(1-\rho_t)\mathbf{d}_t,
\end{equation}
which is Equation~\ref{eq:retrieval_drift}.

The factor $1-\rho_t$ is the full-attention mass omitted by audio KV
retrieval. The vector $\mathbf{d}_t$ measures the discrepancy between
the attention-weighted value representations of the retained and
omitted positions. Consequently, omitting the same attention mass can
have different effects depending on the values associated with that
mass.

Although $\mathbf{d}_t$ cannot be controlled directly using compact
retrieval summaries, the dependence on $\rho_t$ yields a
value-agnostic worst-case bound. Let
\begin{equation}
D_t
=
\max_{i,j\in\mathcal{U}_t}
\left\|
\mathbf{v}_i-\mathbf{v}_j
\right\|_2
\end{equation}
denote the diameter of the value vectors. The retained and omitted
terms in $\mathbf{d}_t$ are convex combinations because
\begin{equation}
\sum_{i\in\mathcal{R}_t}
\frac{\alpha_{t,i}}{\rho_t}
=
1,
\qquad
\sum_{i\notin\mathcal{R}_t}
\frac{\alpha_{t,i}}{1-\rho_t}
=
1.
\end{equation}
Their distance is therefore at most $D_t$, giving
\begin{equation}
\left\|
\widetilde{\mathbf{o}}_t-\mathbf{o}_t
\right\|_2
\leq
(1-\rho_t)D_t,
\label{eq:app_drift_bound}
\end{equation}
which is Equation~\ref{eq:retrieval_drift_bound}.

Under a budget of $K$ retrieved audio positions, the non-audio
contribution to $\rho_t$ is fixed. Selecting the $K$ audio positions
with the largest full-attention weights therefore maximizes $\rho_t$
and minimizes the bound in
Equation~\ref{eq:app_drift_bound}. This full-attention top-$K$
selection provides an oracle retrieval target. Practical methods
cannot compute it without evaluating attention over all audio
positions and must instead approximate audio-token relevance through
compact group-level representations.

%% file: appendix/sections/secC.tex
\section{Implementation Details}
\label{sec:hw_appendix}

For each KV head, \name{} maintains $A$ active clusters, retrieves at
most $B$ audio KV entries, and performs $R$ split--merge exchanges per
decoding step. Although the selected clusters change during
generation, these capacities remain fixed. We therefore use
preallocated buffers and fixed kernel shapes throughout decoding.

\paragraph{Paged decoding.}
The decoding cache uses one-token pages and is stored at each layer as
a contiguous buffer of shape $[P,2,1,H_{\mathrm{kv}},d]$, where $P$ is
the page capacity, the second dimension contains keys and values,
$H_{\mathrm{kv}}$ is the number of KV heads, and $d$ is the head
dimension. One-token pages allow each retrieval slot to contain a
different source position for each KV head.

The logical cache layout is

$$
[\underbrace{\text{retrieved audio}}_{B\text{ slots}}
 \mid \text{non-audio prompt}
 \mid \text{generation}].
$$

Thus, the attention operator always receives $B$ audio slots regardless
of their source positions. We compute exact attention over this cache
using FlashInfer's
\texttt{BatchDecodeWithPagedKVCache} kernel in the \texttt{NHD}
layout. The fixed buffer capacities and launch dimensions also allow
the per-token decoding path to be captured and replayed using CUDA
Graphs.

\paragraph{Selection and KV gathering.}
For each KV head, Stage~1 computes query--centroid scores using
Equation~\ref{eq:node_score} and reduces them across the $G$ associated
query heads. Each audio token inherits the score of its active
ancestor, after which at most $B$ entries are selected. Because
selection is shared within each KV group, at most
$H_{\mathrm{kv}}B$ KV entries are gathered per layer and decoding
step. %For independently selecting baselines, overlapping selections
%are deduplicated before applying the budget-matching protocol in
%Appendix~\ref{app:budget_matching}.
A Triton kernel gathers the selected keys and values directly into
their audio slots in the paged cache. This avoids constructing an
intermediate gathered tensor before attention.

\paragraph{Frontier update.}
We fuse the $R$ split--merge exchanges into a Triton kernel with one
program instance per KV head. Each exchange replaces one active node
with its two children and one active sibling pair with their parent,
thereby preserving the frontier size $A$. Split and merge candidates
are identified through masked maximum and minimum reductions without
sorting. Newly activated children inherit their parent's smoothed
similarity estimate, while a newly activated parent receives the
maximum estimate of its children. The updated frontier and its
metadata are written back to fixed-shape buffers.

\paragraph{Model integration.}
\name{} changes only the selection of audio KVs. Non-audio prompt and
generated-token KVs remain on the model's original cache path, and the
retrieved audio KVs use exact attention. Retrieval is bypassed when
the audio sequence contains no more than $B$ tokens. The method
therefore requires neither retraining nor modification of the
underlying SpeechLLM architecture.

%% file: appendix/sections/secD.tex
\section{Additional Experimental Details}
\label{sec:exp_details}

\subsection{Models and Benchmarks}
\label{app:datasets}

We evaluate Qwen3-Omni-30B-A3B \citep{xu2025qwen3}, Qwen2.5-Omni-3B \citep{Qwen2.5-Omni}, Gemma-4-E4B \citep{team2026gemma}, and Voxtral-Mini-3B \citep{liu2025voxtral}. Within each comparison, all retrieval methods use the same model, benchmark examples, prompts, and decoding settings.

For English ASR, we use the 270 recordings in LibriSpeech-Long (LS-Long) \citep{park2024long}. We construct a multilingual ASR benchmark from Multilingual LibriSpeech \citep{pratap2020mls} by concatenating segments from the same chapter into approximately four-minute recordings in Dutch, German, French, and Spanish, yielding 658 examples. We report corpus word error rate (WER) for both benchmarks. For speech translation, we sample 100 Marco-LongSpeech examples \citep{yang2026longspeech} from each of six languages (Spanish, German, French, Dutch, Italian, and Indonesian) and report corpus chrF. For speech summarization, we use the 545 examples in the first Marco-LongSpeech release partition and report ROUGE-L.

We further evaluate audio understanding on AudioMarathon \citep{he2025audiomarathon}, MMAU-Pro \citep{kumar2026mmau}, and MSP-Podcast \citep{busso2025msppodcastcorpus}. The selected AudioMarathon subsets cover linguistic content, sound events, music, and speaker attributes. MMAU-Pro evaluates general audio understanding, while MSP-Podcast evaluates emotion recognition. To place the short MSP-Podcast utterances in a long-context setting, we pad each utterance with two minutes of silence on each side.

For needle-in-a-haystack evaluation, we construct 120 audio examples of approximately 13--16 minutes. Each contains LibriSpeech-derived background speech and three independently generated seven-digit passcodes, synthesized using Kokoro TTS \citep{hexgrad_2025} and inserted at independently sampled locations. The benchmark is constructed once, and all methods receive identical audio, passcodes, and insertion positions.

\subsection{Retrieval-Budget Matching}
\label{app:budget_matching}

The target budget $B$ is the number of unique audio KV entries
accessed per KV head, layer, and decoding step. Some baselines select
separate subsets for the query heads associated with the same KV head,
so their nominal per-query-head budgets do not directly correspond to
their final KV access.

For method $m$, let
$\mathcal{S}^{(m)}_{t,\ell,h,g}$ be the audio positions selected by
query head $g$ in KV group $h$ at layer $\ell$ and decoding step $t$.
We favorably assume that overlapping selections are deduplicated
before KV access and define
\begin{equation}
\mathcal{U}^{(m)}_{t,\ell,h}
=
\bigcup_{g=1}^{G}\mathcal{S}^{(m)}_{t,\ell,h,g},
\qquad
\bar{B}^{(m)}
=
\mathbb{E}_{x,t,\ell,h}
\left[
\left|\mathcal{U}^{(m)}_{t,\ell,h}\right|
\right].
\label{eq:budget_matching}
\end{equation}

For independently selecting baselines, we calibrate the nominal
per-query-head budget on LibriSpeech-Long so that
$\bar{B}^{(m)}$ is closest to each target
$B\in{128,256,512,1024}$, and reuse the resulting setting across
tasks. Methods that perform one shared selection per KV group,
including UNIQUE and the C2/C3 variants of \name{}, directly target
$B$ and require no calibration. Realized movement for ClusterKV and
Quest/mean remains within approximately $\pm10\%$ of the target on
the recognition benchmarks, with somewhat larger deviations on
translation and summarization. Because of its 16-token page
granularity, Quest cannot reach the 128-KV target except on
Qwen2.5-Omni-3B.

\subsection{Inference and Retrieval Settings}
\label{app:hyperparameter}

All models are evaluated in BF16 with batch size~1 using greedy
decoding and temperature~0. We use fixed benchmark examples,
clustering initialization, and random seeds. The prompt for each task
is held constant across retrieval methods. Retrieval is applied only
to audio KVs; prompt and generated-token KVs remain on the model's
original full-attention path.

Unless otherwise specified, all methods use an average retrieval-unit
granularity of $g=16$. Quest, Quest/mean, and UNIQUE use 16-token
temporal pages, while ClusterKV constructs approximately $N/16$
clusters in key space. \name{} initializes approximately $N/16$
active nodes, with average node sizes of 32, 16, 8, and 4 audio
tokens across its available resolutions.

Clusters are constructed independently for each layer and KV head
using recursive binary $k$-means with eight iterations per split and
random seed~0. Queries and centroids are $\ell_2$-normalized before
cosine-similarity scoring. Unless otherwise specified, \name{} uses
EMA strength $\lambda=0.5$ and performs two paired split--merge
updates per decoding step. The relevance state is initialized to zero.
After a split, both children inherit the parent's state; after a
merge, the parent receives the maximum state of its children.

% \subsection{Evaluation Protocol}
% \label{app:evaluation_protocol}

% Accuracy is reported from one run. Greedy decoding, fixed benchmark
% examples, fixed clustering initialization, and fixed random seeds make
% the evaluation effectively deterministic.

% WER is computed using \texttt{jiwer} after lowercasing, punctuation
% removal, and whitespace normalization. chrF is computed using
% sacreBLEU's corpus chrF.
% ROUGE-L is the mean per-example F-measure computed using \texttt{rouge\_score} with its default tokenizer and Porter stemming. For multiple-choice tasks, accuracy is computed from the letter in the required \texttt{Answer: <letter>} response. Needle-in-a-haystack exact match requires all three passcodes to be recovered correctly; per-digit accuracy is computed over the 21 inserted digits.

%% file: appendix/sections/secE.tex
\section{Additional Experiments}
\label{sec:additional_experiments}

\subsection{Complete Long-Form Speech Results}
\label{app:full_results}

\input{table/main_table_cs16}

Table~\ref{tab:main_table_cs16} expands
Table~\ref{tab:main_table} with intermediate retrieval budgets and
additional temporal-grouping ablations. Quest/mean retains Quest's
fixed temporal partition but replaces its original page summary with
the mean key, separating the effect of temporal grouping from that of
the page representation. We additionally apply shared query-head
competition (C2) and adaptive resolution (C3) to this substrate to test
whether their benefits depend on content-based key-space clustering.

All methods use the default retrieval-unit granularity and are
evaluated under the budget-matching protocol in
Appendix~\ref{app:budget_matching}. Three trends hold across the
complete budget range. First, content-based clustering (C1) generally
outperforms fixed temporal grouping at matched budgets, although both
substrates can fail under the tightest constraints. Second, shared
query-head competition (C2) provides the largest and most consistent
gain, improving both temporal and key-space grouping. Third,
query-adaptive resolution (C3) provides additional gains primarily at
tight budgets, where allocating finer retrieval units around relevant
regions is most valuable. At larger budgets, its marginal effect
narrows and is occasionally slightly negative relative to C2 alone.
Overall, the complete C1\,+\,C2\,+\,C3 configuration is the best or
tied-best retrieval result in 39 of the 48 model--task--budget
settings.

\subsection{Design and Hyperparameter Ablations}
\label{app:ablations}

\input{table/ema_ab}
\input{table/qa}

\paragraph{EMA strength and adaptation rate.}
We evaluate the interaction between the EMA coefficient
$\lambda\in\{0,0.2,0.5,0.8,1.0\}$ and the number of paired
split--merge exchanges $R\in\{1,2,4\}$ per decoding step.
Experiments use Qwen3-Omni and Qwen2.5-Omni-3B at a retrieval budget
of 256 KVs and $g=16$. We sweep the full $\lambda\times R$ grid on
LS-Long and MLS datasets. All other settings remain fixed.

Table~\ref{tab:ema_steps_ablation} shows that
$\lambda=1.0$ is generally the worst setting. In this case,
Equation~\ref{eq:ema_score} never incorporates the current similarity
score, so the refinement state remains fixed and the frontier cannot
follow newly relevant regions.

Away from this degenerate setting, sensitivity is model-dependent.
Qwen3-Omni is relatively insensitive. LS-Long remains within a
$0.6$-WER range ($1.98$--$2.55$), and the default
$\lambda=0.5$, $R=2$ is the best configuration. Qwen2.5-Omni-3B is
more sensitive and benefits from weaker smoothing and faster
adaptation, where LS-Long improves from $11.08$ at the default to $8.75$ at
$\lambda=0.2$, $R=4$, while MLS improves from $31.93$ to $26.55$ at
$\lambda=0.0$.

%We retain a single default of $\lambda=0.5$ and $R=2$ across models for simplicity. It is optimal on the headline Qwen3-Omni model and remains substantially better than the over-smoothed regime on the smaller model.

\begin{figure}[t]
    \centering
    \begin{subfigure}{0.5\linewidth}
        \centering
        \includegraphics[width=0.99\linewidth]
        {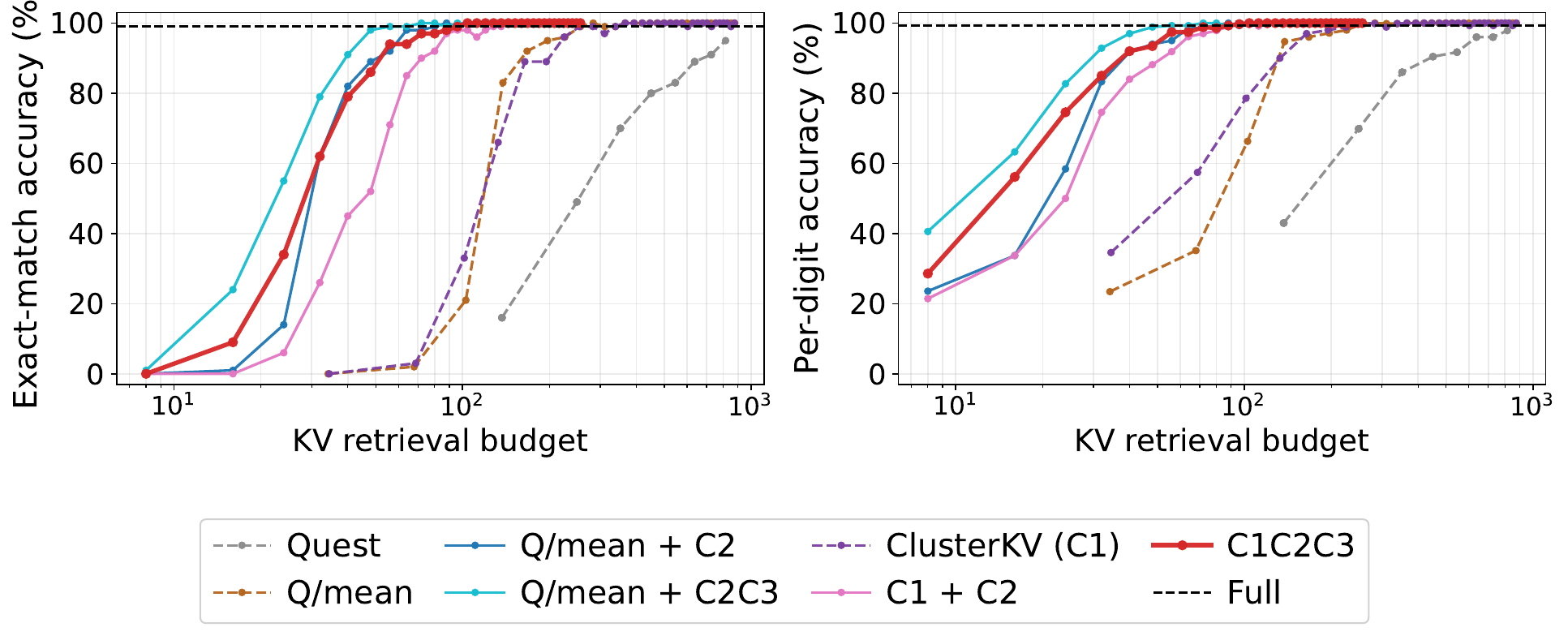}
        \caption{Qwen2.5-Omni-3B}
        \label{fig:niah_qwen25}
    \end{subfigure}%
    \begin{subfigure}{0.5\linewidth}
        \centering
        \includegraphics[width=0.99\linewidth]
        {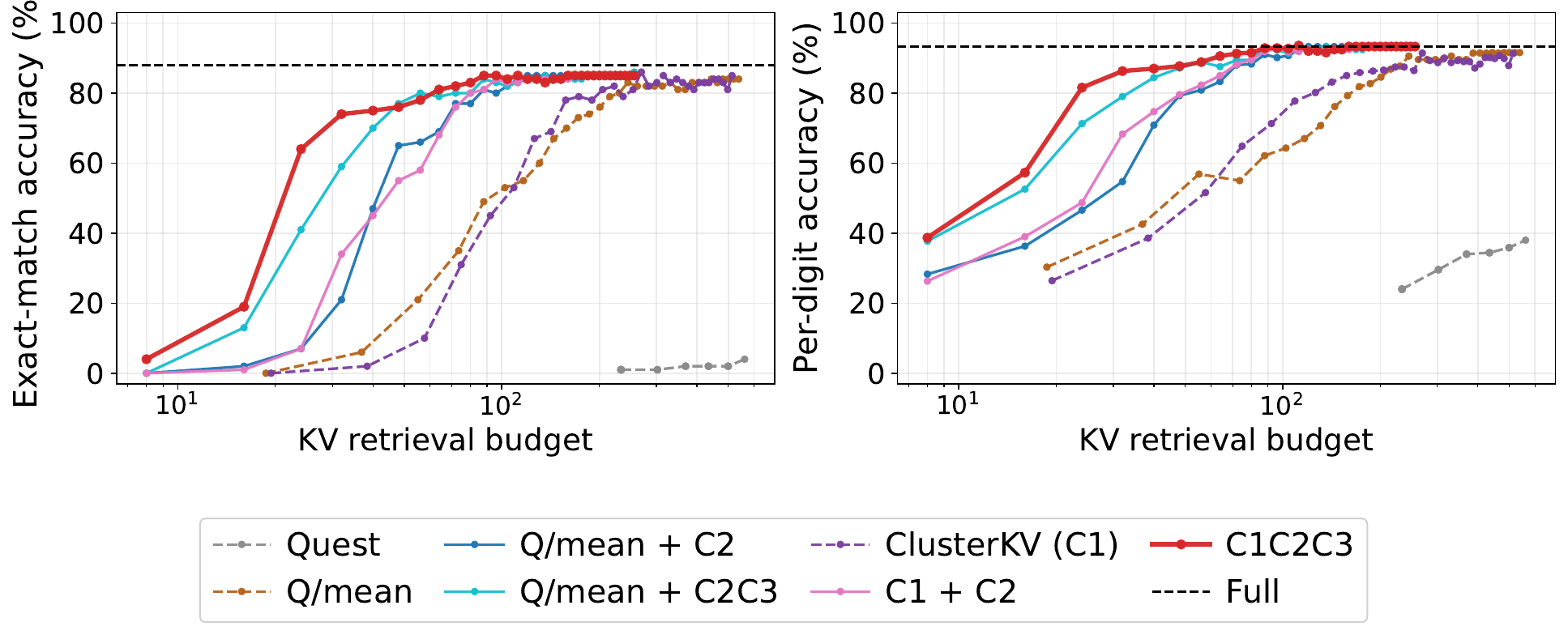}
        \caption{Gemma-4}
        \label{fig:niah_gemma}
    \end{subfigure}
    \begin{subfigure}{0.5\linewidth}
        \centering
        \includegraphics[width=0.99\linewidth]
        {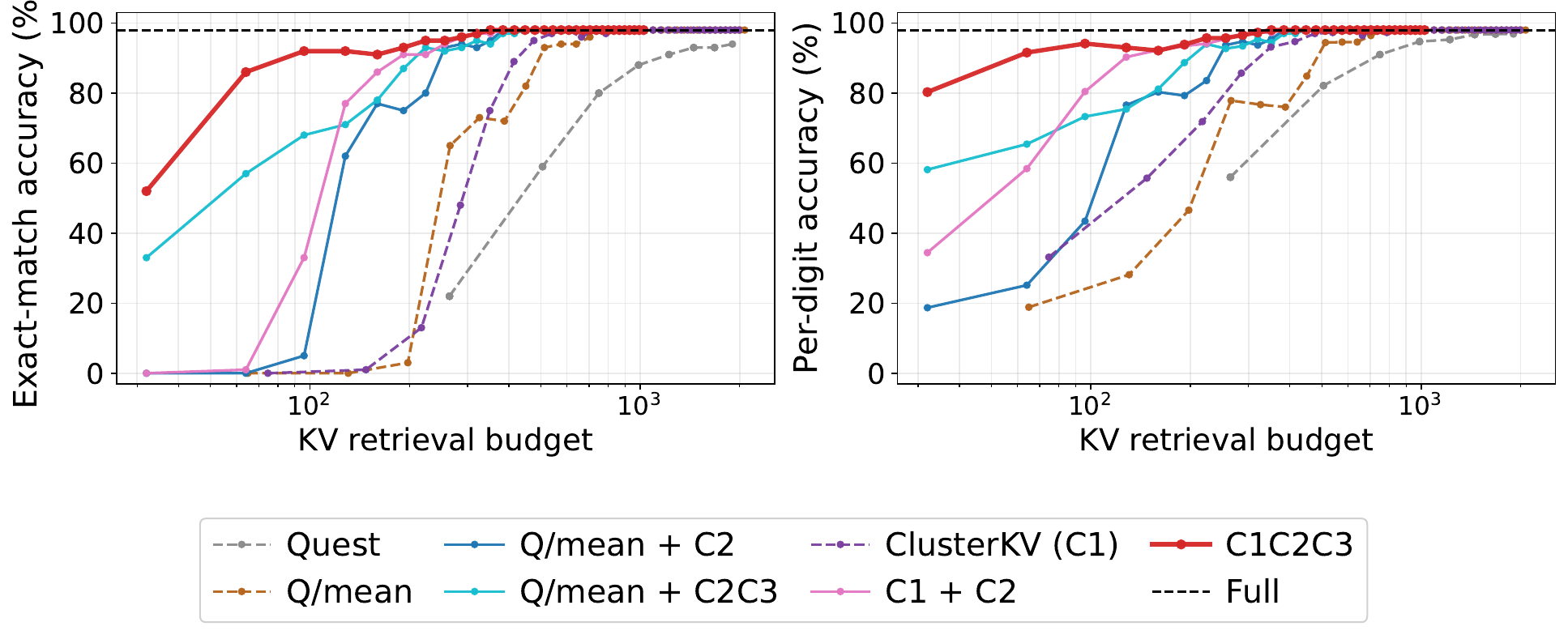}
        \caption{Voxtral-Mini-3B}
        \label{fig:niah_voxtral}
    \end{subfigure}

    \caption{Needle-in-a-haystack performance on Qwen2.5-Omni-3B,
    Gemma-4, and Voxtral-Mini-3B. Each row reports exact-match and
    per-digit accuracy as a function of the audio KV retrieval budget.}
    \label{fig:app_niah}
\end{figure}

\subsection{Additional Audio-Understanding Results}
\label{app:audio_understanding}

Table~\ref{tab:qa} reports additional retrieval results on
AudioMarathon, MMAU-Pro, and MSP-Podcast. Across these benchmarks,
retrieval remains within a few points of full attention at both 128
and 256 KVs in most settings. The component ordering broadly matches
the long-form generation tasks, where shared query-head competition (C2)
provides most of the improvement over temporal grouping and
content-based clustering alone, while C1\,+\,C2\,+\,C3 is the best
or tied-best retrieval configuration in the majority of columns.

The clearest gap appears on Voxtral and the AudioMarathon dataset, where the temporal
baselines fall to approximately $35$--$45\%$, while content-based
clustering with shared competition recovers most of the full-attention
accuracy ($74$--$76\%$ versus $78.8\%$). Adaptive resolution provides
smaller gains on these tasks than on long-form generation, consistent
with their short answer sequences and the correspondingly smaller
opportunity for retrieval errors to compound autoregressively.

\subsection{Additional Needle-in-a-Haystack Results}
\label{sec:app_niah}

Figure~\ref{fig:app_niah} extends the needle-in-a-haystack evaluation
to Qwen2.5-Omni-3B, Gemma-4, and Voxtral. Across models, \name{}
preserves isolated passcode evidence at substantially smaller
retrieval budgets than the temporal and static-clustering baselines.

The budget required to approach each model's full-attention ceiling
varies across architectures. Qwen2.5-Omni-3B reaches approximately
$99$--$100\%$ exact match by roughly $96$--$104$ retrieved KVs,
similar to Qwen3-Omni. Voxtral reaches its approximately $98\%$
full-attention ceiling at a larger budget of roughly 350 KVs. Gemma-4
saturates near $85\%$ exact match by approximately 88 KVs, compared
with an $88\%$ full-attention ceiling, leaving a small residual gap.
Across all three models, \name{} approaches the corresponding ceiling
at substantially smaller budgets than the temporal and
static-clustering baselines, which generally require several hundred
KVs.

%% file: table/main_table_cs16.tex
% requires: \usepackage{booktabs}, \usepackage[table]{xcolor}
\begin{table*}[th]
    \centering
\caption{Long-form speech performance under different audio KV retrieval budgets. We report WER ($\downarrow$) for LibriSpeech-Long (LS-Long) and multilingual long-form ASR (MLS), chrF ($\uparrow$) for speech translation (ST), and ROUGE-L ($\uparrow$) for summarization (Summ). %C1 uses content-based clustering (ClusterKV), C2 adds shared retrieval competition across query heads associated with the same KV head, and C3 adds query-adaptive cluster resolution. All retrieval methods are compared at the same KV budget. 
Best results within each model and budget are shown in bold. Dash (--) for Quest indicates that Quest does not support the target budget.}

    \setlength{\tabcolsep}{2.0pt}   
    \small
    \renewcommand{\arraystretch}{0.85}
\begin{tabular}{l ccc ccc ccc ccc}
\toprule
 & \multicolumn{3}{c}{LS-Long} & \multicolumn{3}{c}{MLS} & \multicolumn{3}{c}{ST} & \multicolumn{3}{c}{Summ} \\
\cmidrule(lr){2-4}\cmidrule(lr){5-7}\cmidrule(lr){8-10}\cmidrule(lr){11-13}
KV Budget & 128 & 256 & 512 & 256 & 512 & 1024 & 256 & 512 & 1024 & 256 & 512 & 1024 \\
\midrule
\multicolumn{13}{c}{\textbf{Qwen3 Omni 30B A3B}} \\
\midrule
Full & \multicolumn{3}{c}{2.02} & \multicolumn{3}{c}{4.83} & \multicolumn{3}{c}{74.75} & \multicolumn{3}{c}{25.52} \\
\midrule
Quest & - & 172.78 & 193.28 & 131.87 & 146.12 & 119.96 & 5.48 & 8.22 & 11.52 & 18.8 & 20.2 & 22.01 \\
Quest/mean & 178.89 & 85.1 & 28.95 & 100.17 & 60.03 & 19.26 & 9.33 & 20.61 & 45.94 & 20.43 & 21.65 & 22.96 \\
\quad +C2 & 70.33 & 20.19 & 3.12 & 46.06 & 12.91 & 5.54 & 20.68 & 46.93 & 62.9 & 21.94 & 23.18 & 24.14 \\
\quad +bias (UNIQUE) & 70.30 & 18.78 & 2.53 & 45.20 & 12.01 & 5.39 & 21.15 & 46.30 & 63.68 & 22.31 & 23.07 & 24.22 \\
\quad +C2+C3 & 35.12 & 8.35 & 2.45 & 28.46 & 9.09 & 5.22 & 35.30 & 53.61 & 64.30 & 22.80 & 23.53 & 24.11 \\
\midrule
C1 (ClusterKV) & 180.94 & 37.44 & 2.81 & 75 & 10.83 & 5.62 & 25.91 & 59.78 & 70.15 & 24.06 & 24.44 & 24.91 \\
\quad +C2 & 24.12 & 2.48 & \textbf{1.97} & 12.9 & \textbf{5.32} & 5.33 & 51.44 & 66.28 & 71.55 & 24.29 & \textbf{24.81} & 24.79 \\
\rowcolor{gray!15} \quad +C2+C3 & \textbf{7.07} & \textbf{1.98} & 1.98 & \textbf{10.05} & 5.39 & \textbf{5.18} & \textbf{55.79} & \textbf{67.05} & \textbf{71.96} & \textbf{24.71} & 24.76 & \textbf{25.29} \\
\midrule
% \multicolumn{13}{c}{\textbf{Qwen2.5 Omni 7B}} \\
% \midrule
% Full & \multicolumn{3}{c}{2.95} & \multicolumn{3}{c}{17.96} & \multicolumn{3}{c}{58.28} & \multicolumn{3}{c}{21.67} \\
% \midrule
% Quest & - & 145.16 & 131.74 & 112.59 & 118.44 & 68.9 & 6.45 & 9.86 & 19.24 & 16.36 & 18.62 & 20.87 \\
% Quest/cent. & 159.87 & 63.7 & 14.65 & 67.47 & 28.44 & 19.18 & 15.6 & 34.62 & 49.02 & 20.01 & 20.13 & 19.97 \\
% \quad +C1 & 75.3 & 29.7 & 7.75 & \textbf{38.95} & \textbf{17.74} & \textbf{15.26} & 22.1 & 40.77 & 53.38 & 20.2 & 20.15 & 19.97 \\
% \quad +C1+C3 & 90 & 49.3 & 11.8 & 58.46 & 26.49 & 16.88 & 20.09 & 33.78 & 47.84 & 20.01 & 20.32 & 20.39 \\
% \midrule
% C2 (ClusterKV) & 103.33 & 34.03 & 5.8 & 46.92 & 22.25 & 19.51 & \textbf{28.27} & \textbf{50.47} & \textbf{56.05} & 20.95 & 20.82 & 21.22 \\
% \quad +C1 & 55.72 & 21.76 & 5.09 & 46.27 & 22.45 & 17.99 & 24.71 & 43.66 & 54.06 & 20.91 & 21.23 & \textbf{21.63} \\
% \rowcolor{gray!15} \quad +C1+C3 & \textbf{50.14} & \textbf{16.67} & \textbf{4.68} & 42.72 & 21.64 & 17.8 & 26.79 & 44.82 & 54.65 & \textbf{20.96} & \textbf{21.27} & 21.61 \\
% \midrule
\multicolumn{13}{c}{\textbf{Qwen2.5 Omni 3B}} \\
\midrule
Full & \multicolumn{3}{c}{2.4} & \multicolumn{3}{c}{14.52} & \multicolumn{3}{c}{42.13} & \multicolumn{3}{c}{24.01} \\
\midrule
Quest & 142.84 & 153.92 & 126.52 & 155.94 & 168.37 & 116.72 & 7.47 & 11.01 & 16.53 & 18.64 & 20.45 & 21.72 \\
Quest/mean & 107.71 & 46.97 & 9.68 & 70.21 & 26.7 & 17.01 & 14.56 & 23.12 & 31.33 & 20.22 & 21.05 & 22.13 \\
\quad +C2 & 55.03 & 24.11 & 3.83 & 43.84 & 16.57 & 14.52 & 17.92 & 28.42 & 36.93 & 20.88 & 22.16 & 23.11 \\
\quad +bias (UNIQUE) & 51.49 & 19.05 & 4.55 & 43.83 & 21.75 & 15.98 & 15.49 & 24.92 & 34.20 & 21.24 & 22.49 & 23.36 \\
\quad +C2+C3 & 63.63 & 31.58 & 6.30 & 58.77 & 28.12 & 16.49 & 19.12 & 26.77 & 35.09 & 21.15 & 22.31 & 23.29 \\
\midrule
C1 (ClusterKV) & 133.13 & 33.11 & 3.91 & 65.32 & 22.01 & 15.61 & 19.32 & 28.7 & 35.01 & 22.68 & 23.07 & 23.97 \\
\quad +C2 & 64.73 & 15.42 & 2.93 & 38.41 & 15.91 & \textbf{13.97} & 19.59 & 31.42 & 38.77 & 22.43 & 23.47 & 24 \\
\rowcolor{gray!15} \quad +C2+C3 & \textbf{40.48} & \textbf{11.08} & \textbf{2.78} & \textbf{31.98} & \textbf{15.72} & 14.31 & \textbf{23.76} & \textbf{33.78} & \textbf{39.35} & \textbf{22.71} & \textbf{23.61} & \textbf{24.11} \\
\midrule
\multicolumn{13}{c}{\textbf{Gemma-4-E4B}} \\
\midrule
Full & \multicolumn{3}{c}{4.24} & \multicolumn{3}{c}{10.26} & \multicolumn{3}{c}{58.97} & \multicolumn{3}{c}{21.64} \\
\midrule
Quest & - & 131.23 & 122.78 & 110.26 & 112.1 & 90.70 & 2.35 & 3.39 & 6.73 & 17.55 & 18.08 & 19.89 \\
Quest/mean & 84.02 & 55.09 & 27.41 & 70.27 & 45.51 & 22.07 & 13.57 & 22.86 & 34.85 & 19.06 & 19.46 & 19.63 \\
\quad +C2 & 39.83 & 17.22 & 7.2 & 32.44 & 17.22 & 11.91 & 26.93 & 37.93 & 48.87 & 18.7 & 19.06 & 19.98 \\
\quad +bias (UNIQUE) & 42.77 & 18.40 & 6.76 & 34.09 & 18.73 & 12.72 & 26.03 & 36.94 & 47.19 & 18.91 & 19.31 & 19.59 \\
\quad +C2+C3 & 24.10 & 11.56 & 5.86 & 24.82 & 16.49 & 12.24 & 32.36 & 41.21 & 49.63 & 18.50 & 19.10 & 19.92 \\
\midrule
C1 (ClusterKV) & 95.72 & 45.95 & 16.73 & 64.05 & 35.15 & 19.26 & 21.28 & 28.74 & 42.17 & \textbf{20.19} & 20.42 & 20.72 \\
\quad +C2 & 22.53 & 7.71 & \textbf{4.91} & 20.2 & 13.44 & 11.19 & 40.12 & 47.91 & 53.72 & 19.71 & \textbf{20.49} & 20.76 \\
\rowcolor{gray!15} \quad +C2+C3 & \textbf{14.63} & \textbf{6.91} & 5.43 & \textbf{18.44} & \textbf{12.90} & \textbf{10.98} & \textbf{42.29} & \textbf{49.06} & \textbf{54.93} & 19.83 & 20.34 & \textbf{20.97} \\
\midrule
\multicolumn{13}{c}{\textbf{Voxtral Mini 3B}} \\
\midrule
Full & \multicolumn{3}{c}{3.39} & \multicolumn{3}{c}{7.29} & \multicolumn{3}{c}{68.43} & \multicolumn{3}{c}{26.94} \\
\midrule
Quest & - & 199.02 & 164.59 & 190.96 & 177.61 & 136.27 & 12.21 & 11.56 & 10.14 & 16.91 & 18.34 & 20.12 \\
Quest/mean & 259.15 & 81.91 & 24.8 & 97.6 & 32.28 & 12.21 & 5.39 & 9.45 & 20.85 & 18.22 & 19.72 & 21.3 \\
\quad +C2 & 99.01 & 49.27 & 11.52 & 56.51 & 16.75 & 9.14 & 5.34 & 14.01 & 29.7 & 18.59 & 20.1 & 22.05 \\
\quad +bias (UNIQUE) & 104.74 & 53.41 & 10.66 & 58.48 & 14.82 & 8.58 & 5.00 & 13.24 & 29.17 & 19.35 & 20.82 & 22.40 \\
\quad +C2+C3 & 43.54 & 19.79 & 8.78 & 24.94 & 13.22 & 8.88 & 10.72 & 19.18 & 30.33 & 19.38 & 20.77 & 22.31 \\
\midrule
C1 (ClusterKV) & 212.96 & 126.8 & 33.64 & 107.58 & 23.77 & 8.21 & 12.42 & 24.73 & 43.74 & 23.29 & 24.6 & 25.74 \\
\quad +C2 & 146.63 & 58.54 & 16.66 & 49.73 & 10.71 & \textbf{7.55} & 13.65 & 28.98 & 44.6 & 23.86 & 25.1 & \textbf{25.92} \\
\rowcolor{gray!15} \quad +C2+C3 & \textbf{41.86} & \textbf{16.12} & \textbf{4.61} & \textbf{14.65} & \textbf{8.17} & 7.61 & \textbf{23.02} & \textbf{36.16} & \textbf{48.84} & \textbf{24.35} & \textbf{25.30} & 25.84 \\
\bottomrule
\end{tabular}
    \label{tab:main_table_cs16}
\end{table*}

%% file: table/ema_ab.tex
\begin{table*}[t]
  \centering
  %\small
  \setlength{\tabcolsep}{5pt}
  \renewcommand{\arraystretch}{1.02}
  \caption{EMA strength $\lambda$ (Eq.~\ref{eq:ema_score}) and adaptation
  rate $R$ (paired split--merge exchanges per decoding step). WER
  ($\downarrow$) at a 256-KV budget with $g{=}16$. The shared default
  configuration ($\lambda{=}0.5$, $R{=}2$) is shaded; the best value per
  model and task block is in \textbf{bold}.}
  \label{tab:ema_steps_ablation}
  \begin{tabular}{l ccccc ccccc}
    \toprule
    & \multicolumn{5}{c}{Qwen3-Omni 30B A3B} & \multicolumn{5}{c}{Qwen2.5-Omni 3B} \\
    \cmidrule(lr){2-6}\cmidrule(lr){7-11}
    EMA $\lambda$ & 0.0 & 0.2 & 0.5 & 0.8 & 1.0 & 0.0 & 0.2 & 0.5 & 0.8 & 1.0 \\
    \midrule
    \multicolumn{11}{l}{\textit{LS-Long}} \\
    $R{=}1$ & 2.31 & 2.27 & 2.31 & 2.22 & 2.46 & 15.79 & 14.88 & 17.10 & 18.81 & 14.31 \\
    $R{=}2$ & 2.04 & 2.21 & \cellcolor{gray!15}\textbf{1.98} & 2.12 & 2.55 & 12.24 & 11.09 & \cellcolor{gray!15}11.08 & 10.31 & 16.89 \\
    $R{=}4$ & 2.07 & 2.07 & 2.12 & 2.07 & 2.55 & 9.76 & \textbf{8.75} & 8.88 & 9.54 & 16.89 \\
    \midrule
    \multicolumn{11}{l}{\textit{MLS}} \\
    $R{=}1$ & 9.96 & 9.89 & 10.37 & 10.89 & 13.85 & 37.26 & 34.35 & 37.11 & 38.49 & 37.38 \\
    $R{=}2$ & 9.84 & 9.85 & \cellcolor{gray!15}10.05 & 10.76 & 13.25 & \textbf{26.55} & 27.33 & \cellcolor{gray!15}31.98 & 32.48 & 37.32 \\
    $R{=}4$ & \textbf{9.62} & 10.05 & 10.29 & 10.20 & 13.23 & 27.46 & 28.61 & 30.08 & 33.04 & 37.31 \\
    \bottomrule
  \end{tabular}
\end{table*}

%% file: table/qa.tex
% requires: \usepackage{booktabs}, \usepackage[table]{xcolor}
% wide: wrap in table*, use \footnotesize and/or \resizebox{\textwidth}{!}{...}
\begin{table*}[]
    \centering
    \caption{Understanding/QA performance under audio KV retrieval budgets, on AudioMarathon (6-subset average accuracy, \%), MMAU-Pro (accuracy, \%), and MSP-Podcast emotion (MSP/acc: accuracy, \%; MSP/F1: macro-F1, \%). Higher is better. Best retrieval result within each model and column is in bold.}%; our full method (C1+C2+C3) is shaded. A dash (--) indicates that Quest cannot reach the 128-KV budget at its page granularity.}
    \setlength{\tabcolsep}{10pt}
        %\small
    %\renewcommand{\arraystretch}{0.85}
\begin{tabular}{l cc cc cc cc}
\toprule
 & \multicolumn{2}{c}{AudioMarathon} & \multicolumn{2}{c}{MMAU-Pro} & \multicolumn{2}{c}{MSP/acc} & \multicolumn{2}{c}{MSP/F1} \\
\cmidrule(lr){2-3}\cmidrule(lr){4-5}\cmidrule(lr){6-7}\cmidrule(lr){8-9}
KV & 128 & 256 & 128 & 256 & 128 & 256 & 128 & 256 \\
\midrule
\multicolumn{9}{c}{\textbf{Qwen3 Omni 30B A3B}} \\
\midrule
Full & \multicolumn{2}{c}{83.3} & \multicolumn{2}{c}{61.8} & \multicolumn{2}{c}{57.9} & \multicolumn{2}{c}{37.2} \\
\midrule
Quest & -- & 79.2 & -- & 57.9 & -- & 54.8 & -- & 33.1 \\
Quest/mean & 78.1 & 80.0 & 57.0 & 59.3 & 53.9 & 56.4 & 30.4 & 34.7 \\
\quad + C2 & 79.9 & 81.3 & 58.9 & 59.9 & 57.3 & 57.8 & 36.1 & 37.0 \\
\quad + C2C3 & 80.5 & 81.7 & 59.5 & 60.2 & 57.4 & 57.9 & 36.1 & 37.0 \\
\midrule
C1 (ClusterKV) & 80.2 & 81.8 & 60.3 & \textbf{61.4} & 54.3 & 56.7 & 31.3 & 35.6 \\
\quad + C2 & 80.9 & 82.1 & 60.7 & 61.2 & 58.0 & \textbf{58.0} & \textbf{37.3} & \textbf{37.3} \\
\rowcolor{gray!15} \quad + C2C3 & \textbf{81.4} & \textbf{82.3} & \textbf{60.9} & 61.3 & \textbf{58.1} & \textbf{58.0} & 37.1 & \textbf{37.3} \\
\midrule
\multicolumn{9}{c}{\textbf{Qwen2.5 Omni 3B}} \\
\midrule
Full & \multicolumn{2}{c}{82.6} & \multicolumn{2}{c}{54.5} & \multicolumn{2}{c}{50.4} & \multicolumn{2}{c}{32.2} \\
\midrule
Quest & 73.5 & 77.9 & 48.7 & 51.2 & 47.7 & 48.4 & 25.2 & 28.7 \\
Quest/mean & 74.0 & 76.4 & 51.3 & 53.1 & 47.7 & 48.2 & 28.4 & 30.4 \\
\quad + C2 & 78.8 & 80.0 & 53.5 & 54.3 & 49.0 & 50.0 & 31.1 & 32.0 \\
\quad + C2C3 & 80.6 & 81.3 & 53.6 & 54.5 & \textbf{49.4} & 50.0 & 31.2 & 31.9 \\
\midrule
C1 (ClusterKV) & 81.3 & 82.2 & 54.1 & 55.6 & 47.0 & 47.5 & 29.1 & 30.9 \\
\quad + C2 & \textbf{82.1} & 82.3 & \textbf{55.8} & \textbf{56.1} & \textbf{49.4} & 50.0 & \textbf{32.2} & 32.0 \\
\rowcolor{gray!15} \quad + C2C3 & \textbf{82.1} & \textbf{82.4} & 55.0 & 55.9 & \textbf{49.4} & \textbf{50.1} & 31.8 & \textbf{32.1} \\
\midrule
\multicolumn{9}{c}{\textbf{Gemma-4-E4B}} \\
\midrule
Full & \multicolumn{2}{c}{78.0} & \multicolumn{2}{c}{46.9} & \multicolumn{2}{c}{50.8} & \multicolumn{2}{c}{28.0} \\
\midrule
Quest & -- & 73.0 & -- & 42.8 & -- & 50.1 & -- & 25.8 \\
Quest/mean & 72.7 & 73.3 & 45.3 & 46.3 & 50.2 & 50.7 & 26.1 & 27.2 \\
\quad + C2 & 74.2 & 75.3 & 46.3 & \textbf{47.1} & \textbf{50.7} & \textbf{50.8} & 27.4 & 27.7 \\
\quad + C2C3 & 74.9 & 75.9 & 46.4 & 46.9 & \textbf{50.7} & \textbf{50.8} & 27.2 & 27.8 \\
\midrule
C1 (ClusterKV) & 74.7 & \textbf{76.7} & 45.8 & 46.1 & 50.6 & \textbf{50.8} & 26.8 & 27.6 \\
\quad + C2 & \textbf{76.4} & \textbf{76.7} & 47.0 & 46.9 & \textbf{50.7} & \textbf{50.8} & \textbf{27.7} & \textbf{27.9} \\
\rowcolor{gray!15} \quad + C2C3 & 76.2 & \textbf{76.7} & \textbf{47.3} & 46.7 & \textbf{50.7} & \textbf{50.8} & \textbf{27.7} & \textbf{27.9} \\
\midrule
\multicolumn{9}{c}{\textbf{Voxtral Mini 3B}} \\
\midrule
Full & \multicolumn{2}{c}{78.8} & \multicolumn{2}{c}{42.0} & \multicolumn{2}{c}{45.9} & \multicolumn{2}{c}{22.6} \\
\midrule
Quest & -- & 44.9 & -- & 35.0 & -- & 43.6 & -- & 10.8 \\
Quest/mean & 38.5 & 38.8 & 35.6 & 38.0 & 43.8 & 45.4 & 12.9 & 19.0 \\
\quad + C2 & 34.3 & 53.5 & 37.4 & 38.8 & 45.5 & 46.1 & 18.2 & 21.5 \\
\quad + C2C3 & 47.9 & 60.0 & 37.4 & 38.8 & 45.6 & 45.9 & 18.7 & 21.1 \\
\midrule
C1 (ClusterKV) & 70.4 & 74.9 & 37.0 & 39.3 & 45.8 & \textbf{46.6} & 17.9 & 22.4 \\
\quad + C2 & 72.6 & 74.9 & \textbf{39.4} & 40.5 & \textbf{46.5} & 46.5 & \textbf{21.0} & \textbf{22.9} \\
\rowcolor{gray!15} \quad + C2C3 & \textbf{74.2} & \textbf{75.6} & 39.1 & \textbf{40.8} & 46.2 & 46.4 & 19.6 & 21.9 \\
\bottomrule
\end{tabular}
    \label{tab:qa}
\end{table*}